%% file: main.tex
\pdfoutput=1

\documentclass[11pt]{article}

\usepackage[preprint]{acl}

\usepackage{times}
\usepackage{latexsym}

\usepackage[T1]{fontenc}

\usepackage[utf8]{inputenc}

\usepackage{microtype}

\usepackage{inconsolata}

\usepackage{graphicx}

\usepackage{multirow}

\newcommand{\ie}{\textit{i.e.}}
\newcommand{\eg}{\textit{e.g.}}

\newcommand{\methodname}{\textsc{Chain-of-Knowledge}}
\newcommand{\dataset}{\textsc{KnowReason}\xspace}

\usepackage{arydshln}
\usepackage{array}
\usepackage{amsmath}
\usepackage{amssymb}
\usepackage{booktabs}
\usepackage{breqn}
\usepackage{colortbl}
\usepackage{etoolbox}
\usepackage{enumitem}
\usepackage{makecell}
\usepackage{multirow}
\usepackage{xspace}
\usepackage{xcolor}
\usepackage{graphicx}
\usepackage{spverbatim}
\usepackage{float}
\usepackage[utf8]{inputenc}
\usepackage{tcolorbox}
\usepackage{multicol}
\usepackage{tabularx}
\usepackage{fontawesome}
\usepackage[linesnumbered,ruled,vlined]{algorithm2e}
\AtBeginEnvironment{spverbatim}{\small}

\definecolor{mygray}{RGB}{226, 226, 226}
\definecolor{myred}{RGB}{252, 142, 142}
\definecolor{mygreen}{RGB}{147, 255, 143}
\definecolor{myblue}{RGB}{144, 155, 255}
\definecolor{myyellow}{RGB}{253, 253, 143}
\definecolor{mypurple}{RGB}{255, 142, 250}
\definecolor{Ground}{RGB}{255,184,55}
\definecolor{Rice}{RGB}{251,248,238}
\definecolor{Dirt}{RGB}{191,169,115}
\definecolor{Pink}{RGB}{226,184,176}
\definecolor{Violet}{RGB}{163,148,170}

\newcolumntype{g}{>{\columncolor{Ground!10}}c}
\newcolumntype{d}{>{\columncolor{Dirt!10}}c}
\newcolumntype{f}{>{\columncolor{Pink!10}}c}
\newcolumntype{v}{>{\columncolor{Violet!10}}c}
\newcolumntype{P}[1]{>{\centering\arraybackslash}p{#1}}

\title{\methodname: Integrating Knowledge Reasoning into Large Language Models by Learning from Knowledge Graphs}

\author {
    Yifei Zhang\textsuperscript{\faMoonO}\thanks{The first two authors contributed equally.}, 
    Xintao Wang\textsuperscript{\faMoonO}\footnotemark[1],\\
    \textbf{
    Jiaqing Liang\textsuperscript{\faMoonO},
    Sirui Xia\textsuperscript{\faMoonO},
    Lida Chen\textsuperscript{\faMoonO},
    Yanghua Xiao \textsuperscript{\faMoonO}}\thanks{Corresponding author.} \\
    \textsuperscript{\faMoonO}Fudan University  \quad \\
    \texttt{\{yifeizhang23, xtwang21\}@m.fudan.edu.cn} \\
    \texttt{\{l.j.q.light,  siruixia39\}@gmail.com} \quad 
    \texttt{shawyh@fudan.edu.cn} \\
}

\begin{document}
\maketitle
\begin{abstract}
\input{Sections/0_Abstract}
\end{abstract}

\input{Sections/1_Introduction}

\input{Sections/2_Background}

\input{Sections/3_Methodology}
\input{Sections/4_Experiments}

\input{Sections/6_Conclusion}

\bibliography{reference}

\input{Sections/Appendix}

\end{document}

%% file: Sections/0_Abstract.tex
Large Language Models (LLMs) have exhibited impressive proficiency in various natural language processing (NLP) tasks, which involve increasingly complex reasoning. 
Knowledge reasoning, a primary type of reasoning, aims at deriving new knowledge from existing one.
While it has been widely studied in the context of knowledge graphs (KGs), knowledge reasoning in LLMs remains underexplored. 
In this paper, we introduce \methodname, a comprehensive  framework for knowledge reasoning, including methodologies for both dataset construction and model learning. 
For dataset construction, we create \dataset via rule mining on KGs. 
For model learning, we observe rule overfitting induced by naive training. Hence, we enhance CoK with a trial-and-error mechanism that 
simulates the human process of internal knowledge exploration. 
We conduct extensive experiments with \dataset.  
Our results show the effectiveness of CoK in refining LLMs in not only knowledge reasoning, but also general reasoning benchmarkms. 

%% file: Sections/1_Introduction.tex
\section{Introduction}

\begin{figure}[t]
    \centering
    \includegraphics[width=1\linewidth]{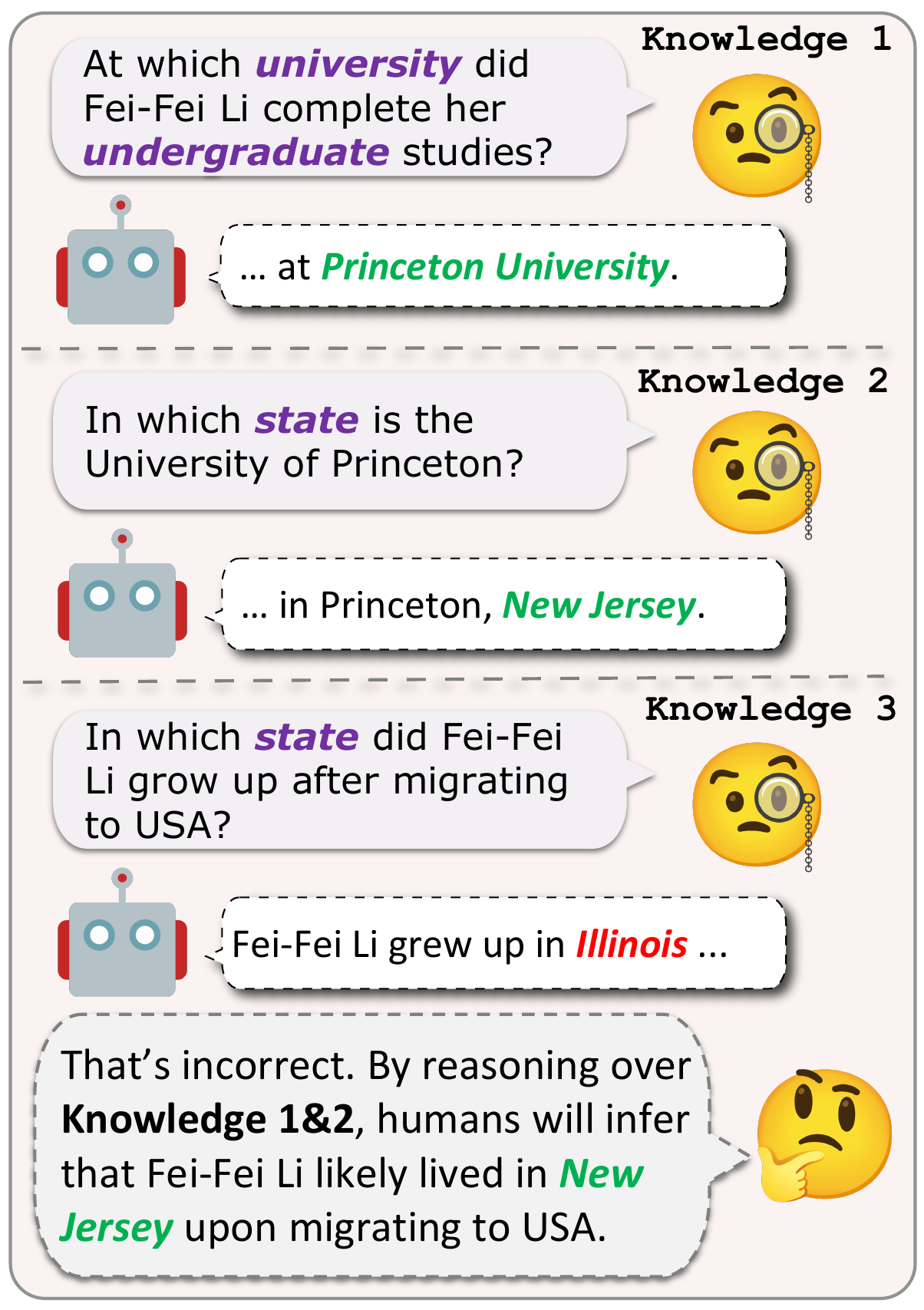}
    \caption{Current LLMs struggle with knowledge reasoning, \ie, combining acquired knowledge to infer new knowledge.}
    \label{fig:1}
\end{figure}

Large Language models (LLMs) have established new state-of-the-arts across a wide range of natural language processing (NLP) tasks~\citep{brown2020language,  bang2023multitask}. 
Increasingly, their impressiveness have  expanded to complex problems challenging reasoning abilities, including arithmetic reasoning~\citep{cobbe2021training}, commonsense reasoning~\citep{talmor2018commonsenseqa}, and symbolic reasoning~\citep{srivastava2022beyond}. 
These reasoning abilities enables LLMs to make informed decisions, solve complex problems, and provide more accurate and relevant responses

Knowledge reasoning represents an indispensable aspect of reasoning, which combines acquired knowledge to derive novel  knowledge~\cite{chen2020review}, as shown in Figure~\ref{fig:1}. 
It shares similarities with commonsense reasoning and symbolic reasoning in its reliance on existing knowledge and logical inference to derive new conclusions.  
Previously, knowledge reasoning has been extensively studied within the context of knowledge graphs (KGs). 
KGs represent fact knowledge in the form of relational triples, \eg, (\textit{Plato}, \textit{author\_of}, \textit{The Republic}).  
Knowledge reasoning over KGs is to harness the existing knowledge to infer and derive novel one, typically by explicitly modeling or implicitly learning compositional rules for relational patterns~\citep{sun2019rotate}. 
This enriches the KGs and supporting downstream tasks such as link prediction~\citep{zhu2021neural} and fact classification~\citep{yao2019kg}. 
Existing methods for KG reasoning could be distinguished into structured-based methods such as TransE~\citep{41864} and description-based methods such as LMKE~\citep{wang2022language}. 
However, knowledge reasoning in LLMs remains significantly underexplored, which could serve as a valuable complement to LLM reasoning.

\begin{figure*}
    \centering
    \includegraphics[width=1\linewidth]{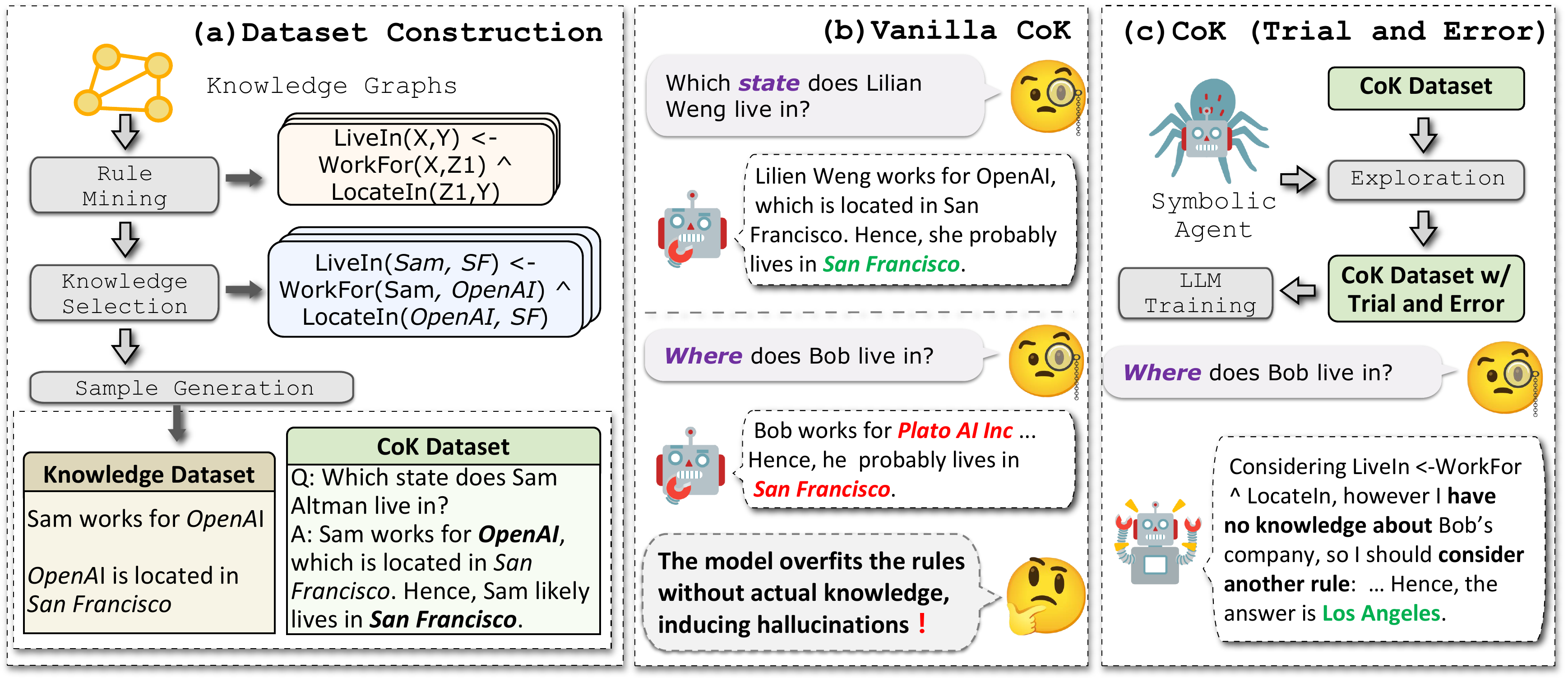}
        \caption{The framework of \methodname (CoK). \textbf{(a) Dataset Construction} includes three steps, \ie, rule mining, knowledge selection and sample generation. This yields a knowledge dataset and CoK dataset. \textbf{(b) Vanilla CoK} trains LLMs in a behavior cloning manner, which may induce rule overfitting and hallucination. \textbf{(c)  CoK (Trial and Error)} is henced proposed, which enables LLMs to simulate humans' internal process of knowledge exploration. }
    \label{fig:2}
\end{figure*}

In this paper, we propose to integrate this knowledge reasoning ability into LLMs, leveraging KGs. 
Specifically, we introduce \methodname~(CoK), a comprehensive learning framework for knowledge reasoning.   
CoK includes methodologies for both dataset construction and model learning.  The \textbf{\textit{dataset construction}} is based on KGs. As illustrated in Figure~\ref{fig:2}, it includes three steps: \textit{1) rule mining} , which mines compositional rules in KGs; \textit{2) knowledge selection}, which identifies interrelated triples matching those rules; and 
\textit{3) sample generation}, which transforms the triples into natural language samples. 
For \textbf{\textit{model learning}}, we observe that training LLMs via behavior cloning often leads to rule overfitting and consequent hallucination. 
Hence, we further enhances CoK with a \textit{trial-and-error} mechanism, which simulates humans' internal process of knowledge exploration  for improved  generalizability.

We conduct extensive experiments to validate the effectiveness of CoK, which covers both anonymized and regular settings. 
In the anonymized settings, we replace entity names to ensure   analysis uninfluenced by data leakage. 
In the regular settings, we showcase the value of CoK for not only real-word knowledge reasoning, but also other reasoning benchmarks. 

The contributions of this paper are mainly summarized as follows:
\begin{itemize}
    \item 
    We introduce the knowledge reasoning task to evaluating and enrich LLMs. 
    Our curated dataset, named \dataset, will be released to facilitate future research in this direction. 
    \item We propose \methodname~(CoK), a comprehensive framework for advancing LLMs' knowledge reasoning ability. 
    CoK provides a detailed method for dataset construction, as well as two  learning methods including behavior cloning and trial-and-error.  
    \item We conduct extensive experiments across various settings, including anonymized ones and regular ones. 
    Our results validate the effectiveness of CoK, with promising generalizability to novel rules and upgraded challenges. 
    Moreover, we showcase the broad utility of CoK via its efficacy in improving other various tasks. 
\end{itemize}

%% file: Sections/2_Background.tex
\section{Related Works}

\paragraph{LLM reasoning}
LLMs have achieved significant success in many NLP tasks and their capabilities have been extended to complex reasoning tasks such as common-sense reasoning~\citep{talmor2018commonsenseqa}, arithmetic reasoning~\citep{cobbe2021training}, and symbolic reasoning~\citep{srivastava2022beyond}. 
It has been observed that LLMs perform poorly on reasoning tasks when using standard prompts~\citep{wei2022chain}. 
To address this issue, Brown et al. proposed few-shot prompting~\citep{brown2020language}, which provides the model with examples of question-answer pairs and has proven effective in reasoning tasks. 
To further enhance performance, \citet{wei2022chain} proposed \textit{Chain-of-Thought} (CoT) prompting~\citep{wei2022chain}, which provides the model with input-output examples that include explicit reasoning steps. 
Different from CoT, Program of Thoughts(PoT) uses language models to express the reasoning process as a program, and then executes the generated programs to derive the answer~\citep{chen2022program}.
Tree-of-Thought (ToT) extends the concept of CoT by incorporating a hierarchical structure into the reasoning process~\citep{yao2024tree}. This approach is particularly useful for tasks requiring complex decision-making and reasoning, where multiple pathways must be evaluated to reach the correct solution. 
Although ToT is effective for decision-making and path selection, it requires access to external information such as context. 
In contrast, our work introduces the CoK framework to enhance the knowledge reasoning abilities of LLMs by utilizing their internal knowledge base.

\paragraph{Knowledge Reasoning over KGs}
Knowledge reasoning is the process of using known knowledge to infer new knowledge~\citep{chen2020review}, which is widely used in knowledge graph completion~\citep{zhang2020relational}. 
Main approaches to knowledge graph reasoning(KGR) can be broadly classified into four main categories: 
\textit{embedding-based reasoning} captures the implicit association between entities and relations through mapping a symbolic representation to a vector space for numerical representation~\citep{41864};
\textit{symbolic-based reasoning} uses logical rules to infer new relationships in knowledge graphs, offering interpretable, human-like reasoning~\citep{galarraga2013amie};
\textit{neural-network-based reasoning} uses neural architectures to predict relationships in knowledge graphs, enabling complex and flexible reasoning~\citep{socher2013reasoning,schlichtkrull2018modeling};
and \textit{mixed reasoning} combines symbolic-based, embedding-based, and neural network-based reasoning to enhance the accuracy and interpretability of knowledge graph reasoning~\citep{xiong2017deeppath,guo2018knowledge}.

%% file: Sections/3_Methodology.tex
\section{Methodology} \label{Methodology}

\subsection{Preliminaries and Task Formulation}
\paragraph{Knowledge Graphs}
KGs store collections of facts as triples, denoted as $G = \{(e, r, e') \mid e, e' \in \mathcal{E}, r \in \mathcal{R}\}$, where $\mathcal{E}$ and $\mathcal{R}$ represent the sets of entities and relations, respectively. 

\paragraph{Atoms and Rules} 
KGs contain compositional rules that can be extracted or modeled to infer new knowledge. 
These rules consist of multiple relational atoms., where an atom can be expressed as $r(X, Y)$, where $r$ is a relation and $X, Y$ are variables for entities. 
A \textit{rule} in the KGs can be expressed by the following formula:
\begin{equation}
r_h(X,Y) \gets r_1(X,Z_1) \wedge \ldots \wedge r_n(Z_{n-1},Y)
\end{equation}
Here, $r_h(X,Y)$ denotes the \textit{head} atom and denotes the \textit{rule head}, while $r_1(X, Z_1) \ldots \wedge r_n(Z_{n-1}, Y)$ denotes the \textit{rule body}.

For example, in the rule $\textit{LiveIn}(X,Y) \gets \textit{WorkFor}(X,Z_1)\wedge \textit{LocateIn}(Z1_Y) \wedge \textit{LiveIn}(X,Y)$ is the rule head and $\textit{WorkFor}(X,Z_1)\wedge \textit{LocateIn}(Z_1,Y)$ is the rule body.

\paragraph{Task Formulation}
Given an atom $r_h(X, Y)$, where $X$ is known and $Y$ is unknown, we seek to determine $Y$. Knowledge reasoning involves identifying an appropriate rule, and then utilize the facts that support the rule body to determine the value of $Y$.

\subsection{Chain-of-Knowledge Data Construction}
\label{cok data}
In this section, we will introduce the idea of our CoK method and how we construct the data.

\paragraph{Rule mining}
In this step, we begin by mining 2-hop rules and then combine them to create 3-hop and 4-hop rules.

To derive rules for data construction from triples in the knowledge graph, we utilize a breadth-first approach to sample 2-hop atoms combinations that connect the head entity to the tail entity. The algorithm we use is shown in Appendix~\ref{appendix:rule mining}
These combinations serve as instances for 2-hop rules. For example, given an atom $r_3(e1,e3)$, we can sample the instance $r_1(e1, e3) \gets r_2(e1, e2) \wedge r_3(e2, e3)$. 
The head and tail of this path correspond to the head entity $e_1$ and the tail entity $e_3$ of the atom respectively.
The corresponding rule is $r_1(X, Y) \gets r_2(X, Z) \wedge r_3(Z, Y)$.

After sampling across the entire knowledge graph, we obtain a series of rule instances. 
First, we count the number of instances corresponding to each rule. 
Rules with fewer than 1000 instances are considered atypical and are removed from the list, serving as the first round of rule filtering.

For a rule $r_1(X, Y) \gets r_2(X, Z) \wedge r_3(Z, Y)$, if the number of instance combinations in the graph that satisfy the \textit{rule body} is $x$, and the number of combinations that also satisfy the \textit{rule head} is $y$, then the \textit{confidence} formula for the rule is $y/x$. 
Using this formula, we calculate the \textit{confidence} for each rule and set a reasonable threshold of 0.6. If the \textit{confidence} is greater than 0.6, we consider the rule to be valid and retain it; otherwise, we discard it. After applying the above steps, we obtain 203 2-hop rules.

We combine 2-hop rules to create longer rules. 
Given two rules, if the rule head of one rule is part of the rule body of another rule, we replace that part with the rule body of the first rule. 
For example, consider the following two rules: 

1) $\textit{BornIn}(X,Y) \gets \textit{HighSchool}(X,Z_1) \wedge \textit{LocateIn}(Z_1,Y)$; 

2) $\textit{CitizenOf}(X,Y) \gets \textit{BornIn}(X,Z_1) \wedge \textit{CityOf}(Z_1,Y)$. 

The head of Rule1, $\textit{BornIn}(X,Y)$, is part of the body of Rule2, so we replace it with the body of Rule1 to form a 3-hop rule: $\textit{CitizenOf}(X,Y) \gets \textit{HighSchool}(X,Z_1) \wedge \textit{LocateIn}(Z_1,Y) \wedge \textit{CityOf}(Z_1,Y)$. 

Using this approach, we generate 159 3-hop rules and 158 4-hop rules.

\paragraph{Knowledge Selection}
In the process of knowledge reasoning, if LLMs do not have access to the facts that support the reasoning path, the reasoning cannot be completed due to the lack of key information. 
On the other hand, if the fact we query is already embedded in the LLMs' internal parameters, the validity of the reasoning cannot be assured due to potential data leakage. 
Therefore, it is essential to select knowledge that is appropriate for CoK data construction.

We construct datasets in both anonymized settings and regular settings, and different knowledge selection processes are applied to the two settings.

For each rule obtained, we first identify all its instances from the knowledge graph.
To prevent the model from overfitting to a particular rule during training, which could lead to path dependency, we ensure a balanced quantity of each rule in the training data. 
We achieve this by sampling an equal number of instances for each rule.
Next, for each instance, we gather the involved facts and use them for knowledge selection.

In the anonymized setting, we replace all entity names in the facts with random, non-existent strings, making all entities new knowledge to the LLMs. 
We then collect all the anonymized facts and use them to generate knowledge data for knowledge injection.

In the regular setting, the entities in the facts represent real-world knowledge. For each instance, we gather the relevant facts and use them to perform knowledge probing on the LLM.
If the LLM knows all the facts supporting the body of the instance but does not know the fact represented by the instance head, we retain this instance for generating samples in the subsequent step.

\paragraph{Sample Generation}
Finally, we apply advanced LLMs~\footnote{In this paper, we use GPT-3.5 for sample generation.} to transform the knowledge into natural language sentences.

For knowledge dataset, we generate a paragraph of natural language description for each entity.
For CoK dataset, we generate a sample for each instance obtained in the previous step . 
For the rule head $r_h(X, Y)$, we prompt advanced LLMs to generate a natural language question.
Given the relationships between entities, if multiple $Y$s correspond to a single $X$, questioning $Y$ will result in multiple answers, complicating evaluation. 
Therefore, we choose the unique entity between $X$ and $Y$ to generate questions.
For the rule body, we combine all the facts to form a reasoning chain that describes the reasoning process from $X$ to $Y$, which will be used as the answer to the question.
The details of sample generation are shown in Appendix~\ref{appendix:sample generation}

Our experiments include both anonymized and regular settings. In each setting, the CoK dataset is used to fine-tune LLMs in a supervised manner.
Conversely, the knowledge dataset is employed exclusively during the continuous pretraining stage in anonymized settings to inject knowledge into LLMs.

\subsection{Chain-of-Knowledge Learning}
\paragraph{Naive Training} 
First, we directly train LLMs on \dataset in a behavior cloning manner.
However, we observe a phenomenon called \textit{rule overfitting}. 
In this situation, the trained models tend to rely on rules encountered during training, even in the absence of supporting facts.

\paragraph{Trial and Error} 
Hence, we introduce a trial-and-error (T\&E) mechanism to CoK learning, which simulates the human process of exploring over our internal knowledge. 

To simulate the human process of knowledge reasoning, we humans initially select a plausible rule and start reasoning based on it when presented with a question.
During this process, if we realize that we lack a crucial fact required by the rule, we switch to an alternative reasoning path instead of continuing without the essential information.

Hence, we integrate the concept of trial and error with our method, incorporating exploration of the LLM's internal knowledge base into the reasoning process. This approach enables LLMs to discern when to apply a rule and when to backtrace it due to a lack of supporting facts, subsequently switching to a more appropriate rule.

We design a symbolic agent to work in conjunction with LLMs to generate exploration path, employing a trial-and-error approach.
For each sample, the symbolic agent first selects a possible rule as a candidate path and then searches for supporting facts for the rule in the internal knowledge base of LLMs. 
If any part of the rule body lacks supporting facts, the process is recorded as an error, and the symbolic agent switches to another rule as the candidate path. 
This process repeats until a reasoning path with sufficient supporting facts is found, leading to the desired result.
The entire exploration process is captured as a data sample, comprising at least one error and the correct reasoning path.
This trial and error process is shown in Algorithm~\ref{algorithm:1}.

\begin{algorithm}
\caption{CoK~(T\&E)}\label{algorithm:1}
\KwData{knowledge graph $G$, rule head $r_h(X,Y)$, large language model $M$}
\KwResult{exploration process $P$}
$t \gets 1$\;
\While{True}{
    \tcp{Select candidate rule}
    $R_t \leftarrow \textit{CandidateRule}(r_h)$\;
    \tcp{Search for supporting facts}
    \For{$fact \in R_t(\text{rule\_body})$}{
        \If{\textbf{not} $\text{IsFactExist}(fact, M)$}{
            $\textit{RecordError}()$\;
            $t \gets t + 1$\;
            \textbf{continue}\;
        }
    }
    \Return{$P$}
}
\end{algorithm}

%% file: Sections/4_Experiments.tex
\section{Experiments}
\subsection{Settings}

\paragraph{Datasets}
We select Wikidata5m~\cite{wang2021kepler} as our data source, which is a million-scale knowledge graph dataset that is aligned with Wikidata, facilitating data processing and usage.

We construct a dataset~\dataset, which includes a knowledge dataset and a CoK dataset. 
The construction method is detailed in Section \ref{Methodology}.

\paragraph{Anonymized Settings and Regular Settings}

We conduct experiments in both the anonymized settings and the regular settings. 

In the \textbf{anonymized settings}, 
we conduct the primarily experiments to study knowledge reasoning in LLMs, avoiding the influence of LLMs' inherent knowledge for this task. 
In these settings, all entity names are replaced with random, non-existent character names, ensuring that the model parameters contain no prior knowledge of these entities. 
Consequently, our training data comprises knowledge dataset used during the continuous pre-training stage, as well as CoK dataset used during the instruction fine-tuning stage.
The knowledge data includes corpus information related to each entity, which injects the necessary prerequisite knowledge for reasoning into the model. 
Meanwhile, the CoK and CoK (T\&E) data serve as the CoK dataset during supervised fine-tuning stage.
The statistics of our training data is showed in Table~\ref{tab:statistic}.

\input{Tables/data_statistic}

In the \textbf{regular settings}, we validate the effectiveness of CoK in real-world scenarios. 
The entities and relationships in the regular settings reflect real-world knowledge.  
Consequently, we fine-tune LLMs with only the CoK dataset to develop the knowledge reasoning ability, without the knowledge injection step.
Besides knowledge reasoning, we 
further evaluate the benefits of CoK learning for LLMs' general reasoning abilities in  downstream tasks.

\paragraph{Evaluation Splits}
To evaluate the knowledge reasoning ability of the model, we designed two test datasets: In-Domain(ID) and Out-of-Domain(OOD) tests.

\textbf{1) In-Domain Tests} The reasoning paths of the samples in ID setting also appear in the training samples. 
A higher score on the ID setting indicates that LLMs can enhance their knowledge reasoning ability by applying the learned rules.

\textbf{2) Out-of-Domain Tests} Unlike the ID setting, the reasoning paths of the samples in the OOD setting do not appear in the training data. 
A higher score on the OOD setting signifies that the knowledge reasoning capability of LLMs effectively generalizes to previously unseen rules.

Additionally, both ID and OOD settings are divided into three subsets, each corresponding to a different rule length.

\paragraph{Models}
We conduct our experiments on Llama3-8B-instruct\cite{llama3modelcard} and Mistral-7b-instruct-v0.2\cite{jiang2023mistral}.

During the fine-tuning stage, we employ full fine-tuning for model training. 
For each training dataset, the model is trained for 4 epochs. After each epoch, it is tested on the evaluation datasets, and the best performance is reported as the result of that training setting.

\paragraph{Methods} \label{exp_methods}
We compare the following four methods in the anonymized settings.
\begin{itemize}
    \item \textbf{Vanilla CoT} In this approach, we prompt the model to answer the question with step-by-step reasoning without any fine-tuning. 
    \item \textbf{In-Context-Learning CoK (ICL-CoK)} In this approach, we provide the model with six examples of question and answer pairs from our CoK learning data.
    \item \textbf{CoK} In this approach, we finetune the model with our CoK data.
    \item \textbf{CoK (T\&E)} In this approach, we finetune the model with our CoK (T\&E) data.
\end{itemize}

\paragraph{Metrics} 
For the knowledge reasoning task, given the rule head $r_h(X,Y)$, we pose a question in natural language to identify $Y$, where $Y$ is the \textit{golden entity} for the question.

We use exact match accuracy as our metric. For each subset of our test data, the evaluation formula is as follows:
\begin{equation}
score(T) = \frac{E}{L(T)}
\end{equation}
where $T$ represents the test dataset, $E$ is the number of samples for which the predicted entity matches the golden answer exactly, and $L(T)$ is the total number of samples in the test dataset $T$.

\subsection{Results in the Anonymized Settings}

\input{Tables/anonymized}
We conduct experiments using each method mentioned in Section \ref{exp_methods}. 
For our CoK and CoK (T\&E) methods, we construct three versions of data for each method using rules of different lengths. 
This approach allows us to explore the relationship between rule length and the model's knowledge reasoning ability.
The results of the experiments are shown in Table~\ref{tab:main_result}.

\paragraph{CoK Effectively Improves LLMs' Knowledge Reasoning Ability.}
Our results show that both CoK and CoK (T\&E) consistently outperform the baselines on all test datasets. 
Notable, given some CoK examples, ICL-CoK generally outperforms vanilla CoT.
However, it still yield relatively low scores, suggesting that LLMs without fine-tuning struggle with knowledge reasoning based on their internal knowledge, despite having key information within their parameters. 

\paragraph{With Trial \& Error, CoK (T\&E) Further Improves Performance in OOD Settings.}
In CoK, the scores for ID dataset are generally higher than those for OOD dataset, demonstrating the phenomenon of \textit{rule overfitting}, where LLMs rely on reasoning paths encountered during training. 
This rule overfitting can lead to hallucinations and a loss of generalization on OOD dataset. 
CoK (T\&E) surpasses CoK on most datasets, with a particularly significant improvement on OOD dataset. 
This suggests that CoK (T\&E) enables LLMs to consider more appropriate rules for questions, rather than blindly applying previously encountered rules.

\paragraph{Learning With Long Rules.}
To study the impact of rule length, we conduct experiments using datasets with varying rule lengths. 
We observe that on the 2-hop and 3-hop test datasets, CoK (T\&E)-3-hop achieves the highest scores, while CoK (T\&E)-4-hop achieves the highest score on the 4-hop test dataset. 
As the rules in the data lengthen, the model's performance does not continuously improve; instead, training with rules having a maximum length of 4-hop actually decreases performance.

To find the reason of this, we calculate the length of rules the model uses in the outputs, the result is shown in Table~\ref{tab:rule_len}.
From the results, the model tend to use longer rules when training with longer rules.
When a shorter path is available, using a longer reasoning path increases the difficulty of reasoning, as each step in the reasoning process requires the model to explore and make decisions.

Training with longer rules helps LLMs learn to use more complex rules in reasoning. 
However, longer rules are not always better. 
Training with them can lead to a tendency to use longer rules during reasoning, even when shorter or simpler paths are available, potentially decreasing performance on simpler tasks.

\input{Tables/output_rule_len}

\paragraph{Error Analysis}
We conduct additional experiments to analyze the reasons for the model’s incorrect predictions in ID and OOD settings.
We categorize the types of errors based on the first incorrect step in the model's reasoning process.
\eg The rule error indicates that the model selects an inappropriate reasoning path, while a fact1 error indicates that the model uses an incorrect $Y$ in the fact $r_1(X, Y)$. 
The results are presented in Table~\ref{tab:error}.

\input{Tables/error_analysis}

The results indicate that on ID dataset, most errors are caused by the model using incorrect facts. This is likely because the reasoning paths in the ID dataset also appear in the training data. 
In contrast, on OOD dataset, more errors are attributable to the reasoning paths selected by the model.
Regarding fact errors, we note that the model frequently hallucinates during the initial stages of entity selection. 
We believe this occurs because, in the later stages of reasoning, the model's knowledge selection scope becomes more restricted as it incorporates additional supporting facts

Specifically, we find that in some error samples, the model selects an appropriate reasoning path and uses correct facts, yet still arrives at an incorrect entity. 
This occurs because for a fact $r_h(X, Y)$, the combination of  $r_h$ and $X$ can correspond to multiple $Y$s. 
Although the model deduces a result different from the ground truth due to using different facts, the reasoning process remain valid.

\subsection{Results in the Regular Settings}

\paragraph{Downstream Tasks with Regular Settings}
In regular settings, we train the model with regular data that utilizes real-world entities for data construction, and further test the model on downstream tasks.

The results of regular settings are shown in Table~\ref{tab:regular}.

\input{Tables/regular_result}

The results from the regular setting validate the conclusions drawn from the anonymized experiments: 
When prompted with CoK examples, ICL-CoK outperforms vanilla CoT on both ID and OOD dataset. 
Both CoK and CoK (T\&E) enhance the LLM's Chain-of-Knowledge capabilities. 
Furthermore, CoK (T\&E) further reduces the LLM's rule dependency, leading to improved performance on OOD dataset.

To further investigate the generalization of CoK, we tested our CoK exploration method on other popular benchmarks. 
The results of the downstream tasks are shown in Table~\ref{tab:downstream_task}. 
The results indicate that CoK (T\&E) outperforms the baseline on three commonsense reasoning benchmarks, suggesting that CoK generalizes well to other reasoning tasks that require various types of knowledge, such as world knowledge.
\input{Tables/downstream_tasks}

%% file: Tables/data_statistic.tex
\setlength\tabcolsep{6pt}
\begin{table*}[!htb]
  \centering
  \small

\begin{tabular}{lcccccc}

\toprule

\multirow{2}{*}{\textbf{Dataset}} & \multirow{2}{*}{\textbf{\#Entity}} & \multirow{2}{*}{\textbf{\#Relation}} & \multirow{2}{*}{\textbf{\#Rule}} & \multirow{2}{*}{\textbf{\#Samples}} & \multicolumn{2}{c}{\textbf{Avg. Sample Hops}} \\
\cmidrule(lr){6-7}
& & & & & \textbf{CoK} & \textbf{CoK(T\&E)} \\

\midrule                    
    
Overall & 6611 & 520 & 644 & 2793 & 2.43 & 5.34 \\
2-hop & 4748 & 203 & 326 & 1993 & 2 & 4.4 \\
3-hop & 978 & 159 & 172 & 400 & 3 & 6.6 \\
4-hop & 885 & 158 & 146 & 400 & 4 & 8.8 \\

\bottomrule
\end{tabular}
\caption{Statistics of the training data in anonymized setting of \dataset dataset}
\label{tab:statistic}
\end{table*}

%% file: Tables/anonymized.tex
\setlength\tabcolsep{7.5pt}
\begin{table*}[ht]
\small
  \centering
\begin{tabular}{lllggggvvvv}
\toprule
\multicolumn{1}{l}{\multirow{2}{*}{\textbf{Model}}} & \multicolumn{1}{l}{\multirow{2}{*}{\textbf{Method}}} & \multirow{2}{*}{\textbf{Rule Length}} & \multicolumn{4}{c}{\textbf{ID}} & \multicolumn{4}{c}{\textbf{OOD}} \\
\cmidrule(lr){4-7} \cmidrule(lr){8-11}
\multicolumn{1}{l}{} & \multicolumn{1}{l}{} &  & \multicolumn{1}{c}{\textbf{2-hop}} & \multicolumn{1}{c}{\textbf{3-hop}} & \multicolumn{1}{c}{\textbf{4-hop}} & \multicolumn{1}{c}{\textbf{all}} & \multicolumn{1}{c}{\textbf{2-hop}} & \multicolumn{1}{c}{\textbf{3-hop}} & \multicolumn{1}{c}{\textbf{4-hop}} & \multicolumn{1}{c}{\textbf{all}} \\
\midrule
\multirow{10}{*}{Mistral-7b} & \multicolumn{1}{l}{Vanilla CoT} & - & 5.47 & 6.97 & 4.48 & 5.64 & 4.98 & 5.47 & 4.98 & 5.14 \\
\cmidrule(lr){2-11}
 & \multirow{3}{*}{ICL-CoK} & 2 & 7.46 & 7.96 & 7.46 & 7.63 & 7.96 & 6.97 & 6.97 & 7.30 \\
 &  & 2\&3-hop & 7.96 & 7.96 & 6.97 & 7.63 & 8.46 & 5.97 & 7.46 & 7.30 \\
 &  & 2\&3\&4-hop & 6.47 & 6.97 & 7.46 & 6.97 & 6.97 & 4.98 & 7.46 & 6.47 \\
 \cmidrule(lr){2-11}
 & \multirow{3}{*}{CoK} & 2 & 11.94 & 14.93 & 12.94 & 13.27 & 16.92 & 7.96 & 5.97 & 10.28 \\
 &  & 2\&3-hop & \underline{15.92} & \underline{19.90} & 16.9 & 17.58 & 16.42 & 8.96 & 14.93 & 13.43 \\
 &  & 2\&3\&4-hop & 13.43 & 15.92 & \underline{21.89} & 17.08 & 12.44 & 16.92 & \underline{22.89} & 17.41 \\
 \cmidrule(lr){2-11}
 & \multirow{3}{*}{CoK(T\&E)} & 2-hop & 11.94 & 8.90 & 9.95 & 10.28 & 14.93 & 11.94 & 11.94 & 12.94 \\
 &  & 2\&3-hop & \textbf{18.41} & \textbf{24.80} & 16.92 & \textbf{20.07} & \textbf{20.90} & \textbf{23.88} & 19.90 & \textbf{21.56} \\
 &  & 2\&3\&4-hop & 14.93 & 17.91 & \textbf{22.89} & \underline{18.57} & \underline{18.41} & \underline{19.90} & \textbf{25.87} & \underline{21.39} \\
\midrule
\multirow{10}{*}{Llama3-8b} & \multicolumn{1}{l}{Vanilla CoT} & - & 4.98 & 6.97 & 8.96 & 6.97 & 5.97 & 8.96 & 5.97 & 6.97 \\
\cmidrule(lr){2-11}
 & \multirow{3}{*}{ICL-CoK} & 2-hop & 8.46 & 6.97 & 6.97 & 7.46 & 7.96 & 6.97 & 6.97 & 7.30 \\
 &  & 2\&3-hop & 7.46 & 7.96 & 6.97 & 7.46 & 7.96 & 7.46 & 5.97 & 7.13 \\
 &  & 2\&3\&4-hop & 7.46 & 8.96 & 8.46 & 8.29 & 6.97 & 8.96 & 8.96 & 8.29 \\
 \cmidrule(lr){2-11}
  \cmidrule(lr){2-11}
 & \multirow{3}{*}{CoK} & 2-hop & \textbf{17.41} & 12.94 & 13.93 & 14.76 & 13.43 & 8.96 & 7.96 & 10.12 \\
 &  & 2\&3-hop & 15.42 & \underline{19.90} & 14.93 & 16.75 & \underline{19.40} & 15.92 & 13.93 & 16.42 \\
 &  & 2\&3\&4-hop & 16.42 & 18.91 & \textbf{21.89} & \underline{19.07} & 10.45 & 17.91 & \underline{19.90} & 16.09 \\
  \cmidrule(lr){2-11}
 & \multirow{3}{*}{CoK(T\&E)} & 2 & 11.94 & 15.92 & 15.92 & 14.59 & 18.91 & 16.92 & 11.94 & 15.92 \\
 &  & 2\&3-hop & 11.94 & \textbf{20.90} & 16.92 & 16.58 & \textbf{21.39} & \textbf{23.88} & 18.91 & \textbf{21.39} \\
 &  & 2\&3\&4-hop & \underline{16.92} & 19.90 & \underline{20.90} & \textbf{19.24} & 12.94 & \underline{21.89} & \textbf{22.89} & \underline{19.24} \\

\bottomrule
\end{tabular}
  \caption{Results(\%) of the experiments on anonymized setting with different methods and different rule length. The best results are \textbf{bolded}, and the second best ones are \underline{underlined}.}
  \label{tab:main_result}
\end{table*}

%% file: Tables/output_rule_len.tex
\begin{table}[t]
\centering
\small
\begin{tabular}{lccc} 
\toprule
\textbf{Training Samples} & \multicolumn{1}{l}{\textbf{2-hop}} & \multicolumn{1}{l}{\textbf{3-hop}} & \multicolumn{1}{l}{\textbf{4-hop}}  \\ 
\midrule
2-hop & 100.0 & 0.0 & 0.0  \\
2\&3-hop & 83.7 & 27.3 & 0.0 \\
2\&3\&4-hop & 47.2 & 24.6 & 28.2 \\
\bottomrule
\end{tabular}
\caption{Proportion(\%) of various rule lengths in the model's output when trained with data of different rule lengths.}
\label{tab:rule_len}
\end{table}

%% file: Tables/error_analysis.tex
\begin{table}[h]
\centering
\begin{tabular}{lccc} 
\toprule
    & \textbf{Rule} & \textbf{Fact1} & \textbf{Fact2}  \\ 
\midrule
ID  & 34.78 & 38.89 & 26.32 \\
OOD & 63.15 & 23.27 & 13.57 \\
\bottomrule
\end{tabular}
\caption{Proportion(\%) of different types of errors in ID and OOD setting across rule, fact1, and fact2.}
\label{tab:error}
\end{table}

%% file: Tables/regular_result.tex
\begin{table}[H]
\centering
\small
\begin{tabular}{llgv} 
\toprule
\textbf{Model} & \textbf{Method} & \multicolumn{1}{c}{\textbf{ID}} & \multicolumn{1}{c}{\textbf{OOD}}  \\ 
\midrule
\multirow{4}{*}{Mistral-7b} & Vanilla CoT  & 0.00  & 0.00 \\
     & ICL-CoK    & 5.50    & 6.20  \\
     & CoK    & 27.00     & 21.89 \\
     & CoK (T\&E) & 21.33  & 22.22 \\
\bottomrule
\end{tabular}

\caption{Performance(\%) of different methods on regular setting.}
\label{tab:regular}
\end{table}

%% file: Tables/downstream_tasks.tex
\setlength\tabcolsep{7.5pt}
\begin{table}[H]
\centering
\resizebox{\linewidth}{!}{

\begin{tabular}{lcccc} 
\toprule
\textbf{Method} & \textbf{CSQA} & \textbf{BBH} & \textbf{ARC-e} & \textbf{ARC-c} \\
\midrule
Mistral-7b & 66.5 & 53.2 & 81.2 & \textbf{72.2} \\
Mistral-7b + CoK (T\&E)  & \textbf{68.1} & \textbf{54.7} & \textbf{82.3} & 69.4 \\
\bottomrule
\end{tabular}
}
\caption{Performance comparison of Mistral-7b and Mistral-7b + CoK (T\&E) on four reasoning benchmarks.}
\label{tab:downstream_task}
\end{table}

%% file: Sections/6_Conclusion.tex
\section{Conclusion}
In this paper, we propose \methodname, a comprehensive learning framework designed to integrate knowledge reasoning abilities into LLMs, encompassing methodologies for both data construction and model learning. 
We construct the \dataset dataset for model training. 
While CoK effectively enhances LLM performance on knowledge reasoning tasks, it may also lead to rule overfitting. 
By employing a trial-and-error approach, CoK (T\&E) addresses this issue and further improves model performance. 
Extensive experiments on two reasoning benchmarks demonstrate the generalization of CoK to other reasoning tasks.

\section*{Limitations}
\paragraph{Evaluation of the Knowledge Reasoning Ability}
Because knowledge reasoning in LLMs remains underexplored in previous studies and no public datasets or benchmarks are suitable for our task, the training and testing of the model's knowledge reasoning ability are conducted using the dataset \dataset that we constructed. 
To address this concern, we strive to ensure the diversity of our dataset.

\paragraph{Data of the Regular Setting is Model-Specific}
In the regular setting, the entities in the data represent real-world knowledge. 
To prevent data leakage, we perform knowledge probing on each model, making the regular setting data model-specific. 
In contrast, for the anonymized setting, we construct data that requires knowledge injection during the continuous pretraining stage but is applicable to all models.

\section*{Ethics Statements}
In this paper, we propose the \methodname~framework, which includes both data construction and a model learning method to enhance the knowledge reasoning ability of LLMs. Our data construction is based on compositional rules from KGs.
First, to ensure the reasonableness of our data, we filter the rules based on their confidence. However, there remains a possibility that some rules may still be unreasonable, leading to flawed samples.
Second, since we utilize advanced LLMs for sample generation, it is inevitable that biases present in the LLMs may influence the knowledge.
To address these ethical concerns, we will optimize the rule mining method and better align the model's output with human cognition.

%% file: Sections/Appendix.tex
\clearpage
\appendix
\section{Details of CoK data construction}
\subsection{Rule minig}\label{appendix:rule mining}
In the rule mining step, we first use a breadth-first search algorithm to find composite rule instances from raw KGs, the algorithm we use is as Algorithm~\ref{algorithm:2}.

\begin{algorithm}
\caption{BFS for rule instances}
\label{algorithm:2}
\For{each triple $(A, r1, B)$}{
    \If{B is in triplets}{
        \For{each $(r2, C)$ for the value of B}{
            \If{A is in triplets}{
                \For{each $(r3, C')$ for the value of A}{
                    \If{$C' == C$}{
                        \Return{combination}
                    }
                }
            }
        }
    }
}
\end{algorithm}

\subsection{Knowledge Selection}
In the anonymized setting, after identifying the supporting facts for the rules, additional processing of the data is required. 
If the head of one instance is part of another instance's body, using this instance for sample generation and model training can lead to data leakage, resulting in an unfair evaluation. 
To address this issue, we separate the head and body parts within the instances. 
By traversing all instances, we create a set for the body facts. 
If an instance's head fact appears in this set, the instance is at risk of data leakage and is therefore discarded.

\subsection{Sample Generation}\label{appendix:sample generation}
\paragraph{Knowledge Dataset}
The knowledge dataset is used to inject knowledge into LLMs exclusively in anonymized settings. It is employed during the continuous pretraining stage and, as such, is presented in the form of a corpus.

For each entity in our knowledge base, we establish a mapping to all related facts. 
Using these facts, we prompt advanced LLMs to generate a descriptive paragraph, encapsulating the knowledge about the entity.
The prompt we use is Prompt~\ref{prompt:1}

\begin{tcolorbox}[title = {Prompt 1: prompt for knowledge corpus generation}, fonttitle=\small]
\small
Rewrite the following sentence as a paragraph describing \{\{entity\}\}, taking care not to change the original meaning of the sentence or lose information:\\
\{\{fact1\}\} \{\{fact2\}\} ... \{\{factn\}\}
\label{prompt:1}
\end{tcolorbox}

To enhance the model's ability to memorize knowledge, we group related facts of each entity into sets of 10 and input them into the LLMs to generate a corpus. 
For each entity, we prompt the LLM to generate four different versions of the knowledge corpus. 
Furthermore, to improve the model's extraction of knowledge from its internal knowledge base, we integrate the corresponding CoK data into the pretraining corpus. 
Thus, each subset of the CoK dataset with different rule lengths has a corresponding knowledge dataset for pretraining.

\paragraph{CoK dataset}
The CoK dataset is used for model learning in the supervise finetuning stage, and has both anonymized and regular setting.
We have 3 steps to generate samples for CoK dataset:

1) \textit{Relation Template Generation}
For each relation, we generate a template sentence which describes the relation of the two entites in the atom.
e.g., having an atom $\textit{CitizenOf}(X,Y)$, the template is \{\{$X$ is a citizen of $Y$\}\}.
The prompt we use to generate the relation templates is as follows:
\begin{tcolorbox}[title = {Prompt 2: prompt for relation template generation}, fonttitle=\small]
\small
You will be given a triple and you should output a sentence which describes the relation between the two entities in the triple, here are some examples:\\
Triple: (<ENT1>, citizen of, <ENT2>)\\
Output: <ENT1> is a citizen of <ENT2>.\\
...\\
Triple: \{\{triple\}\}\\
Output:
\end{tcolorbox}

2)\textit{Question Template Generation}
For each relation in the rule head, we generate a question template for it.
Considering the sufficiency and non-necessity of the rules, and to ensure that each posed question has only one correct answer, we prompt the LLMs to generate a question with possible tone and only question for the unique entity in the atom.
The prompt we use to generate question template is as follows:
\begin{tcolorbox}[title = {Prompt 3: prompt for question template generation}, fonttitle=\small]
\small
You will be provided with a triple, and you should formulate a question that queries the relationship between the two entities. Ensure that the question you generate is in possible tone and has only one correct answer. Here are some examples:\\
Triple: (<ENT1>, citizen of, <ENT2>)\\
Output: Which country may <ENT1> be a citizen of?\\
...\\
Triple: \{\{triple\}\}\\
Output:
\end{tcolorbox}

3)\textit{Sample Generation}
The sample of CoK dataset is in the form of question-answer pairs, so for a sample, we generate question and answer respectively.

For question generation, we substitute the rule head into the corresponding template. For answer generation, we replace each fact in the rule body and rule head, connecting all resulting sentences into a template answer. We then prompt the LLM to polish these sentences into natural language, which serves as the answer for this sample. The prompt we use to generate the natural language answer is as follows:
\begin{tcolorbox}[title = {Prompt 3: prompt for answer generation}, fonttitle=\small]
\small
Rewrite the following sentence into natural language, take care not to change the original meaning or lose information: \{\{sentence\}\}
\end{tcolorbox}

\section{Experiment Settings}
\subsection{Details of Datasets}
\paragraph{Test Dataset of \dataset}
The statistic of the test dataset of \dataset is shown in Table~\ref{tab:testdata}

\input{Tables/testset}

\paragraph{Benchmarks in Downstream Tasks}
\begin{itemize}
    \item \textbf{CommonsenseQA(CSQA)}~\citep{talmor-etal-2019-commonsenseqa}~CommonsenseQA is a new multiple-choice question answering dataset that requires different types of commonsense knowledge to predict the correct answers . It contains 12,102 questions with one correct answer and four distractor answers. 
    \item \textbf{AI2 Reasoning Challenge (ARC)}~\citep{allenai:arc}~ARC is a dataset of 7,787 genuine grade-school level, multiple-choice science questions, assembled to encourage research in advanced question-answering. The dataset is partitioned into a Challenge Set(ARC-c) and an Easy Set(ARC-e).
    \item \textbf{BIG-Bench Hard(BBH)}~\citep{suzgun2022challenging}~BBH is a diverse evaluation suite that focuses on tasks believed to be beyond the capabilities of current language models. It focus on a suite of 23 challenging BIG-Bench tasks for which prior language model evaluations did not outperform the average human-rater. 
\end{itemize}

\subsection{Details of Methods}
\paragraph{Vanilla CoT}
In vanilla CoT, we prompt the model simply with \{\{let's think step by step\}\}.
The prompt we use in as follows:
\begin{tcolorbox}[title = {Prompt 4: prompt for Vanilla CoT}, fonttitle=\small]
\small
\textbf{Instruction:} Knowledge reasoning is the process of using known knowledge to infer new knowledge. You will be given a question of knowledge reasoning task, use you internal knowledge to reason out the answer. Let's think step by step. \\
\textbf{Question:} Which state may Lily live in?
\end{tcolorbox}

\paragraph{ICL-CoK}
In ICL-CoK, we prompt the model with 6 examples from the CoK dataset.
In different rule length setting, we use examples with different rule length, the detail is shown in Table~\ref{tab:icl_cok}
\input{Tables/icl_cok}

The prompt we use is as follows:
\begin{tcolorbox}[title = {Prompt 5: prompt for ICL-CoK}, fonttitle=\small]
\small
\textbf{Instruction:} Knowledge reasoning is the process of using known knowledge to infer new knowledge. You will be given a question of knowledge reasoning task, use you internal knowledge to reason out the answer. \\Here are some examples:\\\{\{Example1\}\}\\...\\\{\{Example6\}\} \\
\textbf{Question:} \{\{question\}\}
\end{tcolorbox}

\paragraph{CoK}
The prompt we use for CoK is as follows:
\begin{tcolorbox}[title = {Prompt 6: prompt for CoK}, fonttitle=\small]
\small
\textbf{Instruction:} Knowledge reasoning is the process of using known knowledge to infer new knowledge. You will be given a question of knowledge reasoning task, use you internal knowledge to reason out the answer. \\
\textbf{Question:} \{\{question\}\}
\end{tcolorbox}

\paragraph{CoK (T\&K)}
The prompt we use for CoK (T\&E) is as follows:
\begin{tcolorbox}[title = {Prompt 7: prompt for CoK (T\&E)}, fonttitle=\small]
\small
\textbf{Instruction:} Knowledge reasoning is the process of using known knowledge to infer new knowledge. You will be given a question of knowledge reasoning task, use you internal knowledge to reason out the answer. If you don't have the knowledge of the supporting fact during reasoning, you should backtrace and change to another path until you can get the answer.\\
\textbf{Question:} \{\{question\}\}
\end{tcolorbox}

\section{Case Study}
\subsection{Examples of Rules}
Table~\ref{tab:rule_example} shows examples of the rules we mine from KGs.
\input{Tables/rule_example}

\subsection{Examples of Knowledge Dataset}
Table~\ref{tab:knowledge_example} shows examples of the knowledge dataset we use in continuous pretraining stage.
\input{Tables/knowledge_example}

\subsection{Examples of CoK Dataset}
Table~\ref{tab:cok_example} shows examples of data in CoK setting.
\input{Tables/cok_example}
Table~\ref{tab:cok_te_example} shows examples of data in CoK (T\&E) setting.
\input{Tables/cok_te_example}

\subsection{Cases of Different Types of Errors}
Table~\ref{tab:case_study} shows different types of errors in error analysis, including rule error, fact1 error and fact2 error.
\input{Tables/case_study}

%% file: Tables/testset.tex
\begin{table*}
\centering
\small
\begin{tabular}{llcccc}
\toprule
\textbf{Setting}              & \textbf{Rule Length} & \textbf{\#Entity} & \textbf{\#Relation} & \textbf{\#Rule} & \textbf{\#Sample}  \\
\midrule
\multirow{3}{*}{ID}  & 2-hop         & 460      & 135        & 147    & 201       \\
                     & 3-hop          & 307      & 96         & 70     & 201       \\
                     & 4-hop          & 360      & 118        & 73     & 201       \\
\midrule
\multirow{3}{*}{OOD} & 2-hop          & 446      & 43         & 28     & 201       \\
                     & 3-hop          & 267      & 66         & 36     & 201       \\
                     & 4-hop          & 353      & 94         & 41     & 201    \\
\bottomrule
\end{tabular}
\caption{Statistics of the test dataset of \dataset}
\label{tab:testdata}
\end{table*}

%% file: Tables/icl_cok.tex
\begin{table}
\centering
\resizebox{\linewidth}{!}{
\begin{tabular}{lccc}
\toprule
\multirow{2}{*}{\textbf{Rule
  Length}} & \multicolumn{3}{c}{\textbf{\# Example}}  \\
  \cmidrule(lr){2-4}
                               & \textbf{2-hop} & \textbf{2\&3-hop} & \textbf{2\&3\&4-hop}           \\
\midrule
2                              & 6     & 0     & 0               \\
3                              & 3     & 3     & 0               \\
4                              & 2     & 2     & 2               \\
\bottomrule
\end{tabular}
}
\caption{\#Examples in ICL-CoK's prompting}
\label{tab:icl_cok}

\end{table}

%% file: Tables/rule_example.tex
\begin{table*}[ht]
\resizebox{\linewidth}{!}{
\begin{tabular}{ll}
\toprule
\textbf{Rule Length} & \textbf{Rules} \\
\midrule
2-hop & \begin{tabular}[c]{@{}l@{}}Country(X,Y)$\gets$PlaceOfBirth(Z,X)$\wedge$CountryOfCitizenship(Z,Y)\\ Contry(X,Y)$\gets$MemeberOfSportsTeam(Z,X)$\wedge$CountryOfCitizenship(Z,Y)\\ CastMember(X,Y)$\gets$OriginalLanguageOfFilm(X,Z)$\wedge$LanguagesSpoken(Y,Z)\end{tabular} \\
\midrule
3-hop & \begin{tabular}[c]{@{}l@{}}Country(X,Y) $\gets$ PlaceOfBirth(Z,X) $\wedge$ ResidentOf(Z,W) $\wedge$ Country(W,Y)\\ Country(X,Y) $\gets$ PlaceOfBirth(Z,X) $\wedge$ EducatedIn(Z,W) $\wedge$ Country(W,Y)\end{tabular} \\
\midrule
4-hop & \begin{tabular}[c]{@{}l@{}}Country(X,Y)$\gets$ PlaceOfBirth(Z,X) $\wedge$ Spouse(Z,W) $\wedge$ PlaceOfBirth(W,V) $\wedge$ CountryOfCitizenship(W,Y)\\ DistantRelative(A,B) $\gets$ PlaceOfBirth(A,X) $\wedge$ ParentOf(Y,A) $\wedge$ CountryOfCitizenship(Y,Z) $\wedge$ Spouse(A,W) $\wedge$ ChildOf(B,W)\end{tabular} \\
\bottomrule
\end{tabular}
}

\caption{Examples of different length of rules we mine}
\label{tab:rule_example}
\end{table*}

%% file: Tables/knowledge_example.tex
\begin{table*}[ht]
\begin{tabularx}{\textwidth}{X}
\toprule
\textbf{Knowledge} \\
\midrule
Excn, a company known for its innovative approach and cutting-edge solutions, has its headquarters situated in Dyxeti. This strategic location serves as the central hub for Excn's operations, allowing the company to efficiently coordinate its various departments and teams. Dyxeti's vibrant business environment and access to top talent make it an ideal setting for Excn to thrive and continue its mission of driving progress and success in the industry. \\
\midrule
Ccmr is the location where the headquarters of Bplx is situated. The presence of Bplx's headquarters in Ccmr is significant as it serves as the central hub for the organization's operations and decision-making. Ccmr, with its strategic position, provides Bplx with easy accessibility to its stakeholders and allows for efficient management of the business. This centralized location in Ccmr ensures effective coordination among various departments and facilitates seamless communication between Bplx and its global network. \\
\midrule
Nryxg is the official language of Gexdzjp, serving as the primary means of communication within the region. It is the designated language used for official documents, government proceedings, and educational instruction, reflecting the cultural and linguistic identity of the people of Gexdzjp. Through the use of Nryxg, individuals in Gexdzjp are able to effectively communicate and connect with one another, fostering a sense of unity and shared understanding among its inhabitants.\\
\bottomrule
\end{tabularx}
\caption{Examples of the knowledge dataset}
\label{tab:knowledge_example}
\end{table*}

%% file: Tables/cok_example.tex
\begin{table*}[ht]
\begin{tabularx}{\linewidth}{>{\raggedright\arraybackslash}p{1.5cm}X>{\raggedright\arraybackslash}p{10cm}X>{\raggedright\arraybackslash}p{3cm}X}
\toprule
\textbf{Rule Length} &
  \textbf{Question} &
  \textbf{Answer} \\
\midrule
2-hop &
  Which country might Anykid be a citizen of? &
  Cckqlvy has Anykid as a part of their team. Cckqlvy is from the country Vevedgta. Therefore, Anykid may be a citizen of Vevedgta. Consequently, Vevedgta is the correct answer. \\
  \midrule
3-hop &
  Which country may Pzdazwi belong to? &
  Mjzk is a citizen of Pzdazwi and is a cast member of Kqts, who is from the country Vevedgta. Therefore, it is possible that Pzdazwi is also from the country Vevedgta. \\
  \midrule
4-hop &
  What may be the original language of Yvag? &
  Brq is fluent in speaking, writing, or signing Owiyxpqf languages. Brq is a actor in Vffdgv, which is produced by Lmljpgzx. Yvag is directed by Lmljpgzx. Therefore, it is possible that the original language of Yvag is Owiyxpqf. \\
  \bottomrule
\end{tabularx}
\caption{Examples of CoK method in CoK dataset}
\label{tab:cok_example}
\end{table*}

%% file: Tables/cok_te_example.tex
\begin{table*}[t]
\begin{tabularx}{\linewidth}{>{\raggedright\arraybackslash}p{1.5cm}X>{\raggedright\arraybackslash}p{10cm}X>{\raggedright\arraybackslash}p{3cm}X}
\toprule
\textbf{Rule Length} &
  \textbf{Question} &
  \textbf{Answer} \\
  \midrule
2-hop &
  To which country does Qzdhsdq belong? &
  To find the answer, we can follow the reasoning path: official_language(Y, X) $\gets$ language_of_work_or_name(A, X), country_of_origin(A, Y). Lzdxfzxw is from Vevedgta, but since we are unsure of Lzdxfzxw's military branch, this path is not applicable. Let's consider a different path: country(X, Y) $\gets$ military_branch(A, X), country_of_citizenship(A, Y). Bmezedu serves in the military branch Qzdhsdq, which is linked to Vevedgta. Therefore, we can conclude that Qzdhsdq is associated with Vevedgta. Hence, the answer is Vevedgta. \\
  \midrule
3-hop &
  What might be the language of Yjuphmb? &
  If a person has a work X and is an author of a book Y, then the language of book Y could be the same as the language of work  X. In this case, Yjuphmb is a work of Kvyps, Kvyps is the author of Tkbz, and the language of Tkbz is Xsbfi. Therefore, it is possible that the language of  Yjuphmb is Xsbfi. \\
  \midrule
4-hop &
  Which country might Liose from? &
  To find the answer, we can use the reasoning path: Country(X,Y) $\gets$ PlaceOfBirth(Z,X) $\wedge$ EducatedIn(Z,W) $\wedge$ Country(W,Y). However, since we are unsure of where Liose was educated, this path is not viable. Let's consider an alternative path: Country(X,Y)$\gets$ PlaceOfBirth(Z,X) $\wedge$ Spouse(Z,W) $\wedge$ PlaceOfBirth(W,V) $\wedge$ LocateIn(W,Y). Since Liose is born is Bhsz and Ari is her spouse, Ari was born in Zhww which locates in Vevegde, it is plausible that Liose is a citizen of Vevegde \\
  \bottomrule
\end{tabularx}
\caption{Examples of CoK (T\&E) method in CoK dataset}
\label{tab:cok_te_example}
\end{table*}

%% file: Tables/case_study.tex
\begin{table*}[]
\begin{tabularx}{\linewidth}{>{\raggedright\arraybackslash}p{1.5cm}X>{\raggedright\arraybackslash}p{8cm}X>{\raggedright\arraybackslash}p{3cm}X>{\raggedright\arraybackslash}p{3cm}X}
\toprule
\textbf{Error Type} & \textbf{Question} & \textbf{Model Output} & \textbf{Explain} \\
\midrule
Rule &
  Can you tell me the country where Nzwscwm is from? &
  Nzwscwm is the country that the author of Wcsa is a citizen of. Wcsa is from the country Vevedgta, so it's possible that Nzwscwm is also from Vevedgta. So the answer is Vevedgta. &
  The reasoning path is not reasonable. \\
  \midrule
Fact1 &
  To which country does Ztgl possibly belong? &
  Ztgl is the head coach of Mfqep. Mfqep is from the country Zxdxcgh. Therefore, it is possible that Ztgl also belongs to the country Zxdxcgh. Thus, Zxdxcgh is the answer. &
  Ztgl is not a coach, the first fact is wrong. \\
  \midrule
Fact2 &
  What is the language Arstkb use? &
  Arstkb has cast member Qoztebgc, who speaks the language Crbzovw. This means that Arstkb use Crbzovw language. Thus, Crbzovw is the answer. &
  Qoztebgc does not speak Crbzovw, the second fact is wrong. \\
  \bottomrule
\end{tabularx}
\caption{Examples of three types of errors}
\label{tab:case_study}
\end{table*}